# Novel Interpretable and Robust Web-based AI Platform for Phishing Email Detection

Abdulla Al-Subaiey
a.al-subaiey.20@abdn.ac.uk
Department of Computing Science, AFG College with the University of Aberdeen, Doha, Qatar

Mohammed Al-Thani
m.al-thani1.20@abdn.ac.uk
Department of Computing Science, AFG College with the University of Aberdeen, Doha, Qatar

Naser Abdullah Alam
naser.abdullah.cse@ulab.edu.bd
Department of Computer Science and Engineering, University of Liberal Arts Bangladesh

Kaniz Fatema Antora
kaniz.fatema.cse@ulab.edu.bd
Department of Computer Science and Engineering, University of Liberal Arts Bangladesh

Amith Khandakar
amitk@qu.edu.qa
Department of Electrical Engineering, College of Engineering, Qatar University

SM Ashfaq Uz Zaman
ashfaquzzaman2@gmail.com
Qatar Emiri Naval Forces, Gulf Arabian, PO BOX 2237, Doha, Qatar

## Abstract

Phishing emails continue to pose a significant threat, causing financial losses and security breaches. This study addresses limitations in existing research, such as reliance on proprietary datasets and lack of real-world application, by proposing a high-performance machine learning model for email classification. Utilizing a comprehensive and largest available public dataset, the model achieves a f1 score of 0.99 and is designed for deployment within relevant applications. Additionally, Explainable AI (XAI) is integrated to enhance user trust. This research offers a practical and highly accurate solution, contributing to the fight against phishing by empowering users with a real-time web-based application for phishing email detection.

**Keywords:** phishing emails, machine learning model, email classification, dataset, explainable ai, user trust, web-based application

## Introduction

The proliferation of online scams and phishing attacks in the digital landscape presents a formidable challenge to cybersecurity efforts worldwide. Phishing, a malicious practice wherein attackers impersonate trusted entities to deceive individuals into divulging sensitive information, remains a pervasive threat. According to recent statistics from Phish Tank, there are over 45 thousand active phishing links, indicating the scale of the issue [1].

Addressing the complexities of phishing demands innovative approaches, among which Machine Learning (ML) and Artificial Intelligence (AI) stand out as promising avenues for strengthening defense mechanisms. ML and AI algorithms, fueled by extensive datasets and pattern recognition capabilities, offer real-time detection of evolving phishing tactics. As highlighted in Cloudflare's 2023 Phishing Threats Report [2] which analyzed data from over 13 billion emails, phishing attacks continue to evolve, with attackers increasingly leveraging deceptive links and identity deception tactics.

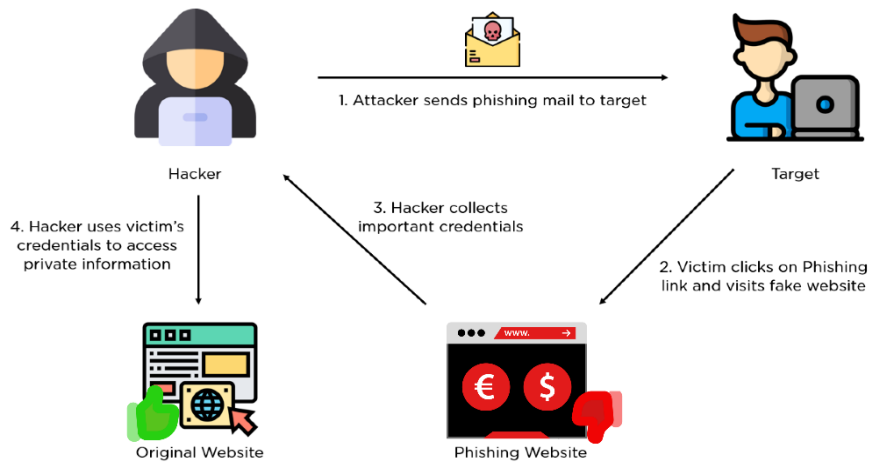

*Figure 1: Phishing using Email* [3]

The urgency of combating phishing is underscored by its significant financial impact. According to the FBI, BEC attacks alone have cost victims worldwide over $50 billion [4]. Based on 4th Quarterly Report of Anti-Phishing Working Group, Inc there have been over 1,077,501 phishing attacks in the fourth quarter of 2023 and 15% for the attacks have been targeted through various webmail services [5]. As the figure below depicts,

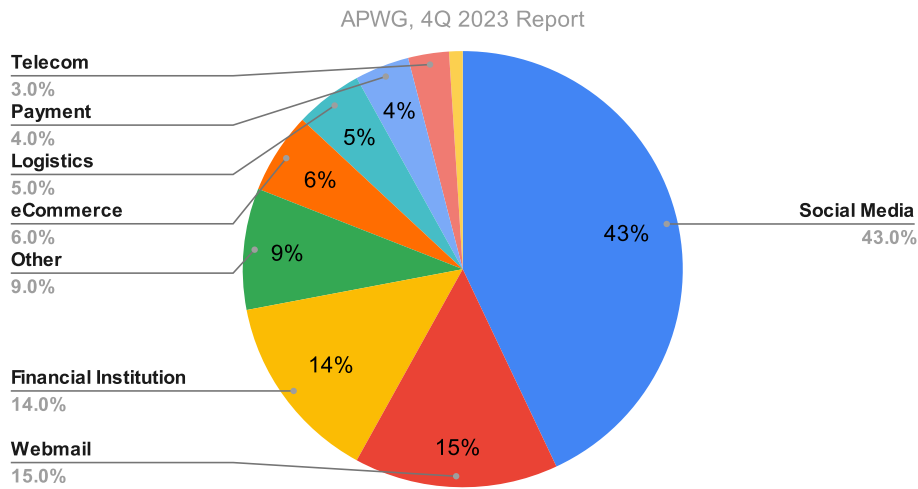

*Figure 2: Most-Targeted Industries for Phishing Attacks. Image generated from APWG 4Q 2023 Report Data*

Moreover, studies indicate that 90% of successful cyber-attacks originate from phishing attempts, therefore making developing robust detection and prevention strategies imperative.

This research aims to contribute to the ongoing efforts in combating phishing by harnessing the power of machine learning algorithms to classify phishing emails effectively. This project aims to develop a web-based application capable of discerning phishing emails from legitimate messages by leveraging insights gleaned from recent phishing trends.

This paper is structured to provide a comprehensive understanding of "Novel Interpretable and Robust Web-based AI Platform for Phishing Email Detection". We begin by introducing the research question and highlighting the significance of the study in Section 1. Section 2 presents a concise review of relevant literature to establish the current understanding of the field. In Section 3, we delve into the methodological approach, detailing the employed dataset, preprocessing techniques, and investigated algorithms. Section 4 focuses on the web deployment of the developed solution, while Section 5 provides a critical discussion of the results obtained. Finally, Section 6 offers concluding remarks, summarizing the key findings, and outlining potential future directions for research.

## Literature Review

While the widespread adoption of artificial intelligence (AI)-based tools has undeniably simplified many aspects of our lives, cybersecurity experts at Kaspersky warn of a potential downside. They believe the rapid growth of AI tools could lead to a double-edged sword: increased accessibility for beneficial applications, but also for malicious actors. This accessibility could fuel the development of more sophisticated cyberattacks. [6].

Recent research work on phishing detection primarily focus on detecting phishing websites from URLs [7], [8], [9], [10], website domain names [11], [12] or website contents [13], [14], [15], [16]. Machine learning (ML) has already shown promise in combating phishing and spam emails. Several studies have extensively reviewed various ML approaches in this field [17], [18]. Another study [19] provided a concise overview of both ML and deep learning strategies, highlighting the critical role of high recall scores for practical implementations.

[20] reviewed studies on Business Email Compromise (BEC) phishing attacks, a sophisticated email scam that impersonates legitimate senders. The analysis of articles published between 2012 and 2022 found several key points: 1) Machine Learning (ML) is a promising approach to detect these evolving attacks, with Decision Tree, Support Vector Machine, and Neural Network being common techniques. 2) Examining email body and header features is crucial for detection, with many studies focusing on both. 3) Future research should explore dynamic feature selection, realistic datasets, integrating Natural Language Processing with deep learning, and combining ML with Explainable AI (XAI) for better detection.

[21] applied a Graph Convolutional Network and various NLP techniques on the CLAIR collection of fraud emails and achieved an accuracy of 98.2%. GCN is, a type of convolutional network that uses graph to represent the relation between entities, it converts the document classification problem into a node classification problem [22], [23]. [24] experimented with various pretrained transformers and machine learning algorithms on publicly available datasets. The experiment achieved a f1 score of 98.66% and accuracy of 98.67% using a fine-tuned BERT transformer. [25] used the Enron email corpus with 6,000 emails (3,000 spam, 3,000 ham) for training and a separate 200-email set (100 spam, 100 ham) for testing. They compared Naive Bayes and SVM with text-based features. SVM achieved higher accuracy (95.5%) than Naive Bayes. Future work will explore richer features and other algorithms. [26] used text tokenization and then implemented a RNN classifier. [27] explored machine learning for spam detection, with Naive Bayes achieving the best results. Bio-inspired algorithms like Genetic Algorithms and Particle Swarm Optimization acted as coaches, fine-tuning the machine learning models to achieve impressive accuracy (even 100% in some cases). This work paves the way for more effective spam filtering by combining machine learning with optimization techniques. [28] proposes ELCADP, a novel ensemble model for lifelong spam classification. It tackles concept drift and catastrophic forgetting through dynamic data partitioning based on drift detection (EDDM). ELCADP achieves superior performance on the

"Enron-Spam" dataset compared to other stream mining methods in terms of accuracy, precision, recall, and F1-score. However, its effectiveness with "virtual concept drift" (new class with same features) remains untested. Future work could explore ELCADP's application to phishing classification, a domain prone to this drift. [29] investigates ensemble methods for spam detection using a multinomial Naive Bayes baseline. Trained on the Kaggle "spam.csv" dataset, the ensemble achieved 98% accuracy. [18] proposes a machine learning approach to detect unsolicited bulk emails (UBEs) using content and behavior-based features. The Random Forest classifier achieved 98.4% accuracy on a ham-spam dataset and 99.4% on a ham-phishing dataset. Future work includes improving robustness and exploring graphical features. [30] proposes an open-source tool for extracting a wide range of features from emails for spam detection. The tool extracts 140 features from the SpamAssassin Public Email Corpus (containing 5051 ham and 1000 spam emails). The performance of four machine learning models (J48 Decision Tree, Multilayer Perceptron, Naive Bayes, Random Forest) was evaluated using these features, achieving very high accuracy (except for Naive Bayes) compared to a previous study. The best performing model was Random Forest. [31] proposes XCSR#, a modification of XCSR (Learning Classifier Systems) to address sentiment analysis and spam detection in social media text. XCSR# tackles sparsity in the data by introducing "don't care" intervals, allowing classifiers to focus on relevant features. Compared to XCSR, XCSR# achieved significant improvement in both tasks (sentiment analysis and spam detection) on datasets like tweets (2034 samples) and SMS spam (5575 samples). However, XCSR# has a potentially higher computational cost due to its more complex condition matching process. [32] proposes a novel email phishing detection method using a hybrid SVM-probabilistic neural network approach. A probabilistic neural network predicts the likelihood of something belonging to a specific category, instead of giving a simple yes/no answer. It achieves improved accuracy 97.5% by dynamically collecting informative features from email text and leveraging content analysis. However, the real-world deployment of the model remains unaddressed. [33] applied RCNN using multilevel vectors and attention mechanisms combined with NLP techniques such as word2vec on the email body and header data and achieved 99% accuracy.

*Table 1: Summary Literature Review*

| Author | Dataset | Method | Result (in %) |
|---|---|---|---|
| [21] | CLAIR collection of fraud email [34]<br>• 3685 spam<br>• 4894 ham | Graph Convolution Network (GCN) and NLP techniques (tokenization, stop word removal) | Accuracy: 98.2<br>False positive rate: 0.015 |
| [24] | Spam Base, and Spam Filter Data (Kaggle),<br>• 3000 spa<br>• 2000 ham | Fine tune BERT transformer | Accuracy: 98.67<br>F1-score: 98.66 |
| [25] | Enron<br>• 3100 spam<br>• 3100 ham | SVM | Accuracy: 95.5<br>Precision: 98.0<br>Recall: 99.0<br>F1-score: 98.5 |
| [26] | SpamAssassin, Enron, Nazario SA-JN<br>• 4572 spam | Preprocess text using tokenization and then applied Recurrent Neural | Accuracy: 98.91<br>Precision: 98.74<br>Recall: 98.53 |

| | | Network (RNN) | F1-score: 98.63 |
| --- | --- | --- | --- |
| | ● 6951 ham<br>En-JN<br>● 9962 spam<br>● 10000 ham | | |
| [27] | Ling, Enron, PUA, SpamAssassin (separately)<br>● 20170 spam<br>● 16545 ham | Genetic Algorithm with SGD (GA-SGD) | Accuracy: 99.21<br>Precision: 98.68<br>Recall: 99.54 |
| [28] | Enron[35]<br>● 17171 spam<br>● 16545 ham | Ensemble based Lifelong Classification using Adjustable Dataset Partitioning (ELCADP) | Accuracy: 95.80<br>Precision: 94.40<br>Recall: 95.80<br>F1-score: 95.10 |
| [29] | Unspecified<br>(Kaggle "spam.csv") | Multinomial Naïve Bayes: using length, stemmer and hyperparameter tuning | Accuracy: 98.00 |
| [18] | SpamAssassin, Nazario<br>● 3051 (2 class)<br>● 3344 (2 class)<br>● 3844 (3 class) | Random forest with fi-based feature selection | Accuracy: 98.40 |
| [30] | SpamAssassin [36]<br>● 1000 spam<br>● 5051 ham | (MLP), Naive Bayes, random forest, and decision tree | Accuracy: 99.30 |
| [31] | Smart home dataset<br>For sentiment analysis<br>● SMS spam (5575 samples)<br>● tweets (2034 samples) | XGBoost, bagged model, and generalized linear model with stepwise feature selection | Accuracy: 91.80 |
| [32] | Private<br>● 404 spam<br>● 1291 ham | SVM with a PNN | Accuracy: 97.5 |
| [33] | Unspecified | RCNN using multilevel vectors and attention mechanisms with Word2Vec | Accuracy: 99.00 |

Literature review exposes limitations in phishing email detection. Most research relies on inaccessible private datasets or small public ones, hindering model generalizability and real-world deployment. Additionally, a gap exists between high-performing models and their practical application. This study addresses these shortcomings by proposing a robust model trained on a comprehensive public dataset and designed for practical use. The research work aims to:
- Address limitations of prior research:
  - Overcome the issue of using proprietary datasets by leveraging a diverse and comprehensive public dataset.
  - Improve generalizability by going beyond small, publicly available datasets used in past studies.
- Focuses on real-world applicability:
  - Design a model for deployment within relevant applications, bridging the gap between theory and practice.

- Aim for a model that can be integrated into websites or applications for real-world use.
- Enhances transparency and trust:
  - Integrates Explainable AI (XAI) to make the model's prediction process more transparent and understandable to users.

The following section details the methodological approach employed in this research endeavor.

# Methodology

The data processing pipeline for this study involved several key stages. First, six spam email datasets were meticulously chosen based on their unique characteristics. These datasets were then merged to create a unified corpus for analysis. Following this, a text preprocessing step was performed, which included tokenization to break down text into meaningful units, and the removal of punctuation and stop words to refine the data. Notably, subject and body text from specific datasets were merged into single "text_combined" columns to streamline further processing. Finally, the preprocessed data from both initial datasets (mdf_1 and mdf_2) were harmonized and integrated based on the "text_combined" column, resulting in a final dataset containing approximately 82,500 emails (42,891 spam and 39,595 legitimate). This prepared dataset was then subjected to feature engineering techniques like TF-IDF and Word2Vec to convert the textual data into a numerical format suitable for machine learning algorithms. Then the dataset was split into training and testing sets and the models were trained and tested. Finally, the best performing model was deployed in a web application.

The image below summarizes the complete process:

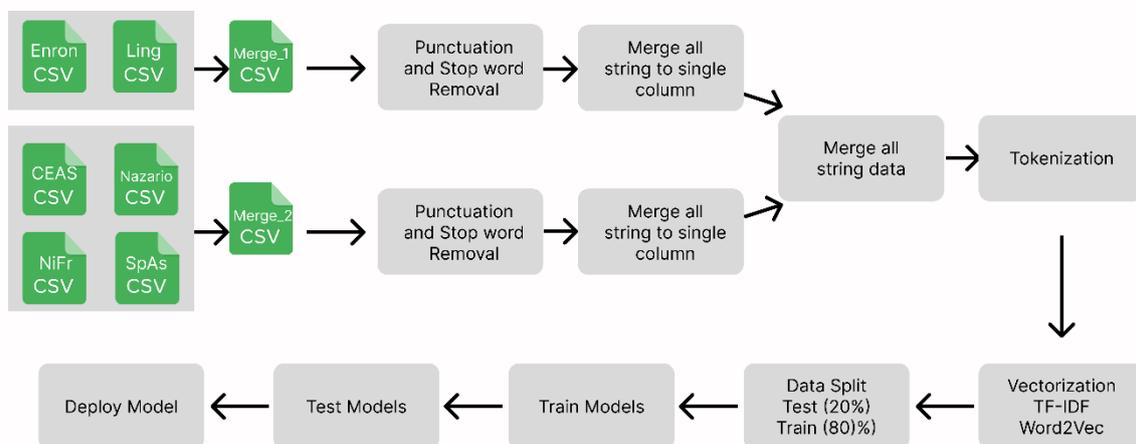

*Figure 3: Methodology*

## 1. Data Collection and Preprocessing

Six widely used spam email datasets were carefully selected based on their unique attributes. These datasets underwent a merging process to create a unified dataset for analysis. Among these, the Enron and Ling datasets contained three crucial columns: subject, body, and label, amalgamated into a singular data frame denoted as mdf_1. The CEAS, Nazario, Nigerian

Fraud, and SpamAssassin datasets included sender, receiver, subject, body, date, and label columns, all integrated into another data frame labeled mdf_2.

## 2. Text Processing

### 2.1. Tokenization and Text Cleaning

Utilizing advanced Natural Language Processing (NLP) techniques, the textual data underwent tokenization to break down words into meaningful units. Further, punctuation marks and stop words were removed to refine the text for subsequent analysis.

### 2.2. Text Combination

The subject and body columns from mdf_1 were merged into a unified column labeled 'text_combined'. Similarly, the sender, date, subject, and body columns from mdf_2 were merged into a single column named 'text_combined'. This consolidation was executed to streamline further processing stages.

## 3. Data Integration and Preparation

The datasets originating from mdf_1 and mdf_2 was harmonized into a cohesive dataset. The data was merged in the 'text_combined' column. This integrated dataset comprised 42,891 spam emails and 39,595 legitimate (ham) emails, totaling almost 82,500.

## 4. Feature Engineering: Text Preprocessing and Vectorization

The textual data within the email corpus requires preprocessing and transformation into a numerical format suitable for machine learning algorithms. This section details the feature engineering process, specifically focusing on vectorization techniques employed: Term Frequency-Inverse Document Frequency (TF-IDF) and Word2Vec.

### Term Frequency-Inverse Document Frequency (TF-IDF):

TF-IDF is a statistical weighting scheme that evaluates the importance of a word within a document relative to the entire document collection (corpus) [37]. It considers two crucial factors:
1. Term Frequency (TF): Measures the frequency of a word appearing within a specific document. A higher TF indicates the word is more prominent within that document.
2. Inverse Document Frequency (IDF): Captures the global importance of a word across the entire corpus. Words that appear frequently across all documents will have a lower IDF score, signifying less discriminatory power. Conversely, words unique to a few documents will have a higher IDF, indicating their potential relevance for distinguishing specific content.

The mathematical formulation of TF-IDF for a word $w$ in a document $d$ within a corpus D is:

$$TF - IDF(w, d) \;=\; TF(w, d) \;\times\; IDF(w, D) \qquad \ldots (1)$$

- $TF(w, d)$ is often calculated as the number of times word $w$ appears in document $d$ divided by the total number of words in document $d$.
- $IDF(w, D)$ is calculated using the logarithm of the total number of documents in the corpus ($|D|$) divided by the number of documents containing word $w$ ($df(w)$).

TF-IDF addresses the shortcomings of simply using word frequency. By incorporating the IDF component, it reduces the weight of frequently used words that appear frequently across all documents and emphasizes terms specific to a particular document or category. In the context of phishing email detection, TF-IDF can highlight unique keywords or phrases commonly used in phishing attempts, enhancing the model's ability to identify such emails.

### Word2Vec:

Word2Vec is a neural network-based technique that represents words as numerical vectors. It leverages the idea that words with similar meanings tend to appear in similar contexts within a corpus. The training process allows Word2Vec to learn these semantic relationships and embed words into a vector space, where words with closer meanings will have more similar vector representations.[38].

### Vectorization Process:

In this study, both TF-IDF and Word2Vec were employed to convert the preprocessed text within the 'text_combined' column into numerical representations. This transformation allows machine learning algorithms to analyze the textual features and identify patterns indicative of phishing emails.

## 5. Model Development and Evaluation

The dataset was split into an 80-20 train test ratio. After the split, there were 65988 training samples and 16498 testing samples out of 82486.

| Training | Testing | Total |
|----------|---------|-------|
| 65988    | 16498   | 82486 |

The thorough review presented by [17], [18], [19], [20] suggest various techniques and algorithms implemented to address this challenge. From the vast list of techniques, we have chosen to equip the project with the following models to develop the system.

1. Support Vector Classifier (SVC)
   In this work, we employ a Support Vector Machine (SVM) for classification using a linear kernel. SVMs find a hyperplane that maximizes the margin between classes, effectively separating the data. The linear kernel facilitates efficient computation in high dimensions while maintaining interpretability. For binary classification, SVC with a linear kernel aims to find a straight line (hyperplane) in high dimensional space that best separates the two classes. This line maximizes the margin between the closest data points of each class (support vectors). New data points are then classified based on which side of the hyperplane they fall on. To ensure reproducibility, a random state of 42 is set for all computations.

2. Multinomial Naive Bayes Classifier (Multinomial NB)
   We utilize a Multinomial Naive Bayes (MNB) classifier, a probabilistic model suited for discrete features like word counts in text data. MNB calculates class probabilities based on feature independence and predicts the class with the highest likelihood. This approach offers efficiency and interpretability for text classification tasks.

3. Random Forest Classifier
   We employ a Random Forest Classifier consisting of 100 decision trees for improved accuracy and reduced overfitting. Each tree acts independently, analyzing the data using a random selection of features at each branching point. When a new data point arrives, all the trees vote for a class based on their individual learned rules. Finally, the majority vote from the entire forest determines the final classification for the data point. This ensemble approach helps prevent overfitting and can potentially lead to better accuracy compared to a single decision tree. Setting the random state at 42 ensures reproducibility of the model.

The models were chosen due to the following reasoning,
1. Strong Performance
2. Computational Efficiency
3. Interpretability
4. Advantages over other models

The thorough study [39] shows both MNB and SVC have achieved high accuracy (over 95%) in similar context. Furthermore, the research work states the simplicity compared to deep learning make these algorithms more interpretable. [25], [39] suggest that SVM is slightly more robust compared to MNB however, MNB is known for its efficiency in handling large datasets. In our literature review we have also seen significant contribution of Random Forest Classifiers therefore we included all of the algorithms stated above.

Evaluation Metrics

Comprehensive evaluation of the performance of the developed models in classifying phishing emails, various metrics were employed. These metrics provide insights into different aspects of a model's effectiveness:

**Accuracy:**
Accuracy is the most intuitive metric, representing the overall proportion of correctly classified emails. It is calculated as the number of true positives (correctly classified phishing emails) and true negatives (correctly classified legitimate emails) divided by the total number of emails:

$$Accuracy = \frac{(True\ Positives\ +\ True\ Negatives)}{Total\ Emails} \quad \ldots (2)$$

However, accuracy can be misleading, particularly in imbalanced datasets where one class (e.g., phishing emails) might be significantly smaller than the other (legitimate emails). In such cases, a model could achieve high accuracy simply by predicting the majority class.

**Precision:**

Precision focuses on the positive predictive value, indicating the proportion of predicted phishing emails that are true positives. It is calculated as the number of true positives divided by the total number of emails predicted as phishing emails (true positives + false positives):

$$Precision = \frac{True\ Positives}{(True\ Positives\ +\ False\ Positives)} \quad \ldots (3)$$

A high precision value suggests the model effectively avoids classifying legitimate emails as phishing emails (low false positive rate).

**Recall:**
Recall, also known as sensitivity, emphasizes the model's ability to identify all relevant phishing emails. It is calculated as the number of true positives divided by the total number of actual phishing emails (true positives + false negatives):

$$Recall = \frac{True\ Positives}{(True\ Positives\ +\ False\ Negatives)} \quad \ldots (4)$$

A high recall value indicates the model successfully captures most of the phishing emails within the dataset (low false negative rate).

**F1-Score:**
The F1-score is a harmonic mean that combines the strengths of precision and recall, providing a more balanced view of the model's performance. It is calculated as:

$$F1\_Score = 2 \times \frac{Precision * Recall}{(Precision\ +\ Recall)} \quad \ldots (5)$$

An F1-score closer to 1 signifies a well-balanced model with high precision and recall.
By evaluating these metrics in conjunction, we gain a comprehensive understanding of the model's effectiveness in detecting phishing emails. A high accuracy score along with balanced precision and recall indicates a robust model for real-world application.

## 6. Model Interpretability with LIME

While achieving high performance is crucial, understanding the rationale behind a model's predictions is equally valuable. This section details the application of Local Interpretable Model-Agnostic Explanations (LIME) to gain insights into the decision-making process of our machine learning model for phishing email detection.
LIME is a technique for explaining individual predictions made by any complex machine learning model. It operates by approximating the behavior of the original model around a specific data point (an email instance in our case) with a simpler, interpretable model, often a linear model. This local explanation model focuses on features within the data point that hold the most influence on the original model's prediction.

In the context of our research, LIME is employed to analyze how the model classifies an email as phishing. LIME takes a pre-classified email instance (e.g., an email flagged as phishing by the model) and generates an explanation for that classification. This explanation highlights the features within the email content that most significantly contributed to the model's decision. These features might include specific words, phrases, or patterns within the email text.

## 7. Model Deployment

Following its superior performance on evaluation metrics, the chosen model is exported and integrated into a Flask web application. This deployment strategy facilitates real-time spam email classification, enabling users to interact with the model through a user-friendly web interface.

# Results

## Model Results

The table below concisely summarizes the experiment results. In the dataset column, first, we have the number of samples and then the class. For example, in the first row, 28457[1] means there are 28457 spam emails, and 21403[0] means 21403 ham emails. In the table, [1] and [0] refer to spam and ham classes, respectively.

*Table 2: Experiment Results*

| Model | Preprocessing | Dataset | Accuracy | Precision | Recall | F1-score |
|---|---|---|---|---|---|---|
| svm_0.713 | word2vec | 42891[1] 39595[0] | 0.713 | 0.7 | 0.72 | 0.71 |
| svm_0.821 | word2vec | 42891[1] 39595[0] | 0.821 | 0.82 | 0.81 | 0.82 |
| rf_0.838 | word2vec | 42891[1] 39595[0] | 0.838 | 0.83 | 0.84 | 0.83 |
| mnb_985 | tf-idf | 28457[1] 21403[0] | 0.985 | 0.98 | 0.99 | 0.99 |
| svm_994 | tf-idf | 28457[1] 21403[0] | 0.994 | 0.99 | 0.99 | 0.99 |
| rf_988-url | tf-idf | 28457[1] 21403[0] | 0.988 | 0.98 | 0.99 | 0.99 |
| mnb_984-url | tf-idf | 42891[1], 39595[0] | 0.978 | 0.97 | 0.99 | 0.98 |
| svm_991-url | tf-idf | 42891[1], 39595[0] | 0.991 | 0.99 | 0.99 | 0.99 |
| rf_984 | tf-idf | 42891[1], 39595[0] | 0.984 | 0.98 | 0.99 | 0.98 |
| **svm_991 (proposed)** | **tf-idf** | **42891[1], 39595[0]** | **0.991** | **0.99** | **0.99** | **0.99** |
| mnb_978 | tf-idf | 42891[1], 39595[0] | 0.978 | 0.97 | 0.99 | 0.98 |
| rf_984 | tf-idf | 42891[1], 39595[0] | 0.984 | 0.98 | 0.99 | 0.98 |

The best-performing model, **SVM** with **TF-IDF** preprocessing on the merged dataset, achieved 99.1% accuracy, 99% precision, 99% recall, and f1-score 99.

*Table 3: Result Comparison with Literature*

| Author | Dataset | Method | Result |
|---|---|---|---|
| [21] | CLAIR collection of fraud email [34]<br>● 3685 spam<br>● 4894 ham | Graph Convolution Network (GCN) and NLP techniques (tokenization, stop word removal) | Accuracy: 98.2<br>False positive rate: 0.015 |
| [24] | Spam Base, and Spam Filter Data (Kaggle),<br>● 3000 spa<br>● 2000 ham | Fine tune BERT transformer | Accuracy: 98.67<br>F1-score: 98.66 |
| [25] | Enron<br>● 3100 spam<br>● 3100 ham | SVM | Accuracy: 95.5<br>Precision: 98.0<br>Recall: 99.0<br>F1-score: 98.5 |
| [26] | SpamAssassin, Enron, Nazario<br>SA-JN<br>● 4572 spam<br>● 6951 ham<br>En-JN<br>● 9962 spam<br>● 10000 ham | Preprocess text using tokenization and then applied Recurrent Neural Network (RNN) | Accuracy: 98.91<br>Precision: 98.74<br>Recall: 98.53<br>F1-score: 98.63 |
| [27] | Ling, Enron, PUA, SpamAssassin (separately)<br>● 20170 spam<br>● 16545 ham | Genetic Algorithm with SGD (GA-SGD) | Accuracy: 99.21<br>Precision: 98.68<br>Recall: 99.54 |
| [28] | Enron<br>● 17171 spam<br>● 16545 ham | Ensemble based Lifelong Classification using Adjustable Dataset Partitioning (ELCADP) | Accuracy: 95.80<br>Precision: 94.40<br>Recall: 95.80<br>F1-score: 95.10 |
| [29] | Unspecified (Kaggle "spam.csv") | Multinomial Naïve Bayes: using length, stemmer and hyperparameter tuning | Accuracy: 98.00 |
| [18] | SpamAssassin, Nazario<br>● 3051 (2 class)<br>● 3344 (2 class)<br>● 3844 (3 class) | Random forest with fi-based feature selection | Accuracy: 98.40 |
| [30] | SpamAssassin<br>● 1000 spam<br>● 5051 ham | (MLP), Naive Bayes, random forest, and decision tree | Accuracy: 99.30 |

| Ref | Dataset | Model | Results |
|---|---|---|---|
| [31] | Smart home dataset<br>For sentiment analysis<br>• SMS spam (5575 samples)<br>• tweets (2034 samples) | XGBoost, bagged model, and generalized linear model with stepwise feature selection | Accuracy: 91.80 |
| [32] | Private<br>• 404 spam<br>• 1291 ham | SVM with a PNN | Accuracy: 97.5 |
| [33] | Unspecified | RCNN using multilevel vectors and attention mechanisms with Word2Vec | Accuracy: 99.00 |
| **Proposed Model** | Enron, Ling, CEAS, SpamAssassin, Nazario, Nigerian Fraud<br>• 42891 spam<br>• 39595 ham | SVC using linear kernel, tf-idf vectorizer, and lime visualizer | Accuracy: 99.10<br>Precision: 99.00<br>Recall: 99.00<br>F1-score: 99.00 |

The following section describes the deployment of the best performing model within a web application. We demonstrate the application's functionality using real-world data and showcase prediction visualization techniques like LIME for interpretability.

# Web-based Platform

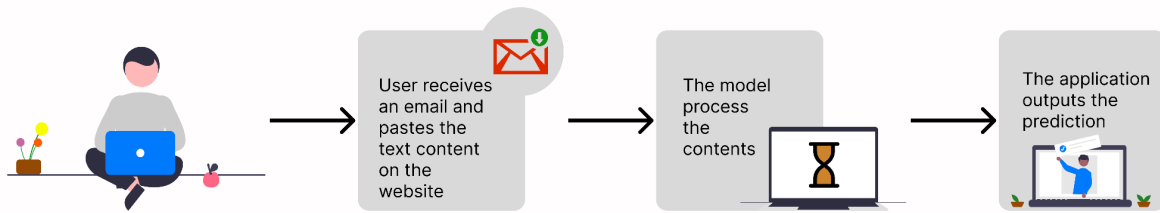

*Figure 4: Web-application process flow*

The figure above depicts how the user shall interact with the web application. The moment the user receives a suspicious email, the user can copy and paste the text contents (sender address, subject, body text) directly to the website and submit the form. The model embedded within the website will instantly send a prediction weather the email contents are spam or safe.

The figures below, are a demonstration of the application and the model in real time. The model can predict unseen and real-life emails. The user pastes the email text contents as can be seen below in the form. Then when the verify button is pressed a request is sent to the model to predict the contents. The web application processes the text data and converts it into a vector and then passes the data to the model. The model predicts based on the rich set of features it was trained on and provides a result. The website is live at https://phishingdetection.onrender.com/.

*Figure 5: Real world email before prediction*

*Figure 6: Real world email after prediction*

Feature Importance Analysis and Prediction Visualization using LIME

The LIME visualization presents a snippet of a phishing email and analyzes the likelihood of it being spam [40]. The email subject is "Personal Assistant Opportunity - Dr. Sheldon Cooper" and it is addressed to an unknown applicant. The email body offers a work-from-home assistant position with a competitive salary. It specifies tasks such as errands, communication, and some academic responsibilities. The sender requests the applicant to send their CV, phone number, and a scanned copy of their passport for verification to an email address (shel.cooper@caltech.edu) that appears to be affiliated with California Institute of Technology (Caltech). The email further asks the applicant to fill out and scan a job application form. The demo email was generated based on the recent events from University of Aberdeen [41].

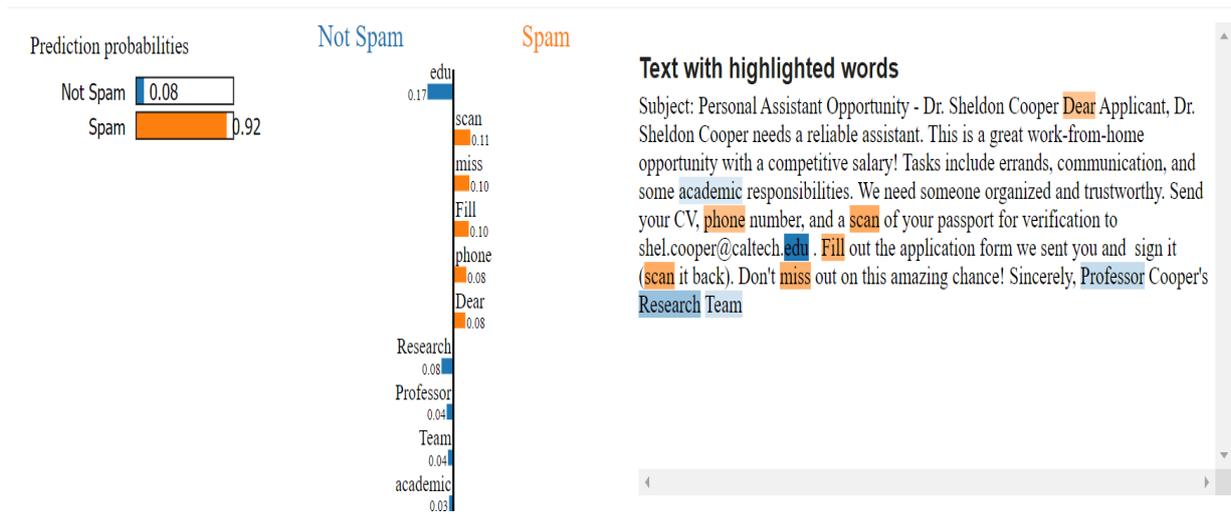

*Figure 7: LIME visualization after prediction*

The LIME visualization divides the prediction into two sections: Not Spam and Spam. The "Not Spam" section has a probability score of 0.08, while "Spam" has a score of 0.92, indicating a high likelihood of the email being spam.

Highlighted words within the email body are colored red, signifying their contribution to the spam classification. These words include "scan" (0.11), "miss" (0.10), "Fill" (0.10), "phone" (0.08), and "Dear" (0.08). Conversely, the word "edu" within the sender's email address is colored blue and has a weight of 0.03, indicating that it contributes slightly to the email being classified as "Not Spam" because "edu" addresses are commonly associated with educational institutions.

In essence, the LIME visualization demonstrates how the model identifies specific keywords typically found in phishing emails, such as requests for personal documents (passport scans) and urgency ("Don't miss out on this amazing chance!"), to classify the email as spam.

The following section delves into the various aspects of this research endeavor. We will explore how specific results informed key decisions made throughout the project.

# Discussion

## Feature Engineering and Model Selection

This study investigated the influence of feature preprocessing techniques on model performance for phishing email detection. Two common vectorizers, word2vec and tf-idf, were compared. Our experiments revealed that tf-idf achieved superior results, with an F1 score of 0.99 compared to the maximum F1 score of 0.83 obtained using word2vec. Based on this finding, tf-idf was employed for subsequent model training.

We further conducted feature ablation to identify the most informative features for the classification task. Initially, all available features (sender email, receiver email, date, subject, body, URL) were included in the model. We hypothesized that the receiver email address would have minimal influence on phishing email detection, as spammers often employ spoofing techniques. This hypothesis was validated, as removing the receiver email feature resulted in no significant performance change. Similarly, the URL column, containing binary data (indicating presence or absence of a URL), appeared to have minimal predictive power, and was excluded without impacting model performance.

A more impactful observation was the significant improvement in model performance achieved by merging all textual features (sender email, date, subject, and body) into a single column. This merged feature yielded a notable increase in F1 score, from 0.71 to 0.82. This suggests that combining textual information provides a richer representation for the model compared to using isolated features. The merged feature captures contextual relationships between these textual elements, potentially aiding the model in identifying phishing patterns. Finally, SVM's ability to consider word relationships, unlike Naive Bayes' assumption of word independence, contributes to its better performance.

### Enhancing Transparency with Explainable AI (XAI)

To promote interpretability and gain insights into the model's decision-making process, we integrated Explainable AI (XAI) techniques. This approach aims to visualize the features that most significantly influence the model's predictions. In the context of phishing email detection, XAI can be used to identify specific words or phrases that contribute the most to classifying an email as spam. This information can be valuable for improving future iterations of the model and potentially for user education, highlighting the red flags commonly used in phishing attempts.

# Conclusion

In summary, the growing threat of phishing attacks highlights the importance of robust cybersecurity defenses. Numerous statistics and reports have demonstrated that phishing remains a pervasive and constantly evolving hazard, posing significant financial and security risks to individuals and organizations worldwide. Leveraging the power of Machine Learning (ML) and Artificial Intelligence (AI) offers a promising avenue for enhancing detection and prevention strategies against phishing attacks.

This research project makes a meaningful contribution to the ongoing efforts to combat phishing by developing a high-performing machine-learning model that effectively classifies phishing emails. The proposed model achieved remarkable accuracy and precision rates by leveraging insights from recent phishing trends and drawing on a comprehensive dataset merged from multiple sources. Notably, SVM with TF-IDF preprocessing achieved 99.1% accuracy and commendable precision, recall, and f1-score metrics.

Moreover, deploying the model within a web-based application represents a significant step towards practical implementation and real-world applicability. The application improves the efficacy of phishing email detection by enabling users to interact with the model in real time. It empowers individuals and organizations to mitigate the risks associated with phishing attacks.

Given the limitations observed in existing literature, which include reliance on proprietary datasets and a lack of real-world deployment, this study emphasizes the importance of scalable, generalizable models deployed in practical scenarios. Going forward, continued research and collaboration in phishing email detection are critical to staying ahead of evolving threats and safeguarding digital ecosystems against malicious actors.

## Data Availability

The processed dataset used in this study can be made available upon a reasonable request to the corresponding author.